\documentclass[11pt,a4paper]{article}
\usepackage[hyperref]{acl2019}
\usepackage{times}
\usepackage{multirow}
\usepackage{enumitem}
\usepackage{latexsym}
\usepackage{graphicx}
\usepackage{csquotes}
\usepackage{physics}
\usepackage{textcomp}
\usepackage{subfig}
\usepackage{caption}

\usepackage{lipsum}     
\usepackage{xargs}      
\usepackage[colorinlistoftodos,prependcaption,textsize=tiny]{todonotes}
\newcommandx{\unsure}[2][1=]{\todo[linecolor=red,backgroundcolor=red!25,bordercolor=red,#1]{#2}}
\newcommandx{\change}[2][1=]{\todo[linecolor=blue,backgroundcolor=blue!25,bordercolor=blue,#1]{#2}}
\newcommandx{\info}[2][1=]{\todo[linecolor=green,backgroundcolor=green!25,bordercolor=green,#1]{#2}}
\newcommandx{\improvement}[2][1=]{\todo[linecolor=purple,backgroundcolor=purple!25,bordercolor=purple,#1]{#2}}

\usepackage{url}

\aclfinalcopy 
\def\aclpaperid{***} 

\title{On the impressive performance of \\ randomly weighted encoders in summarization tasks}

\author{Jonathan Pilault$^{1,2,3,}$\Thanks{ Equal contribution, order determined by coin flip},   Jaehong Park$^{1,*}$, \textbf{Christopher Pal}$^{1,2,3,4}$ \\
$^{1}$Element AI, $^{2}$Montreal Institute for Learning Algorithms, \\$^{3}$Ecole Polytechnique de Montreal, $^{4}$Canada CIFAR AI Chair \\
{\texttt{$^{1}$\{jonathan.pilault, jaehong.park\}@elementai.com}}\\
}

\date{}

\begin{document}
\maketitle
\begin{abstract}
In this work, we investigate the performance of \emph{untrained randomly initialized} encoders in a general class of sequence to sequence models and compare their performance with that of \emph{fully-trained} encoders on the task of abstractive summarization. We hypothesize that random projections of an input text have enough representational power to encode the hierarchical structure of sentences and semantics of documents. Using a trained decoder to produce abstractive text summaries, we empirically demonstrate that architectures with untrained randomly initialized encoders perform competitively with respect to the equivalent architectures with fully-trained encoders. We further find that the capacity of the encoder not only improves overall model generalization but also closes the performance gap between untrained randomly initialized and full-trained encoders. To our knowledge, it is the first time that general sequence to sequence models with attention are assessed for trained and randomly projected representations on abstractive summarization.
  
\end{abstract}

\section{Introduction}
Recent state-of-the-art Natural Language Processing (NLP) models operate directly on raw input text, thus sidestepping typical prepossessing steps in classical NLP that use hand-crafted features \cite{young2018recent}. It is typically assumed that such engineered features are not needed since critical parts of language are modeled directly by encoded word and sentence representations in Deep Neural Networks (DNN). For instance, researchers have attempted to evaluate the ability of recurrent neural networks (RNN) to represent lexical, structural or compositional semantics \cite{DBLP:journals/corr/LinzenDG16, DBLP:journals/corr/abs-1711-10203, DBLP:journals/corr/abs-1711-00350}, and study morphological learning in machine translation \cite{DBLP:journals/corr/BelinkovDDSG17, Dalvi2017UnderstandingAI}. Various diagnostic methods have been proposed to analyze the linguistic properties that a fixed length vector can hold \cite{W16-2524, DBLP:journals/corr/AdiKBLG16, DBLP:journals/corr/KielaCJN17}.

Nevertheless, relatively little is still known about the exact properties that can be learned and encoded in sentence or document representations from training. While general linguistic structures has been shown to be important in NLP \citep{strubell-mccallum-2018-syntax}, knowing whether this information comes from the architectural bias or the trained weights can be meaningful in designing better performing models. Recently, it was demonstrated that randomly parameterized combinations of pre-trained word embeddings often have comparable performance to fully-trained sentence embeddings \cite{notraining}. Such experiments question the gains of trained modern sentence embeddings over random methods. By showing that random encoders perform close to state-of-the-art sentence embeddings, \citet{notraining} challenged the assumption that sentence embeddings are greatly improved from training an encoder.

As a follow-up to \citet{notraining}, we generalize their approaches to more complex sequence-to-sequence (seq2seq) learning, particularly on abstractive text summarization. We investigate various aspects of random encoders using a Hierarchical Recurrent Encoder Decoder (HRED) architecture that either has (1) an untrained, randomly initialized encoders or (2) a fully trained encoders. In this work, we seek to answer three main questions: (i) How effective are untrained randomly initialized hierarchical RNNs in capturing document structure and semantics? (ii) Are untrained encoders close in performance to trained encoders on a challenging task such as long-text summarization tasks? (iii) How does the capacity of encoder or decoder affect the quality of generated summaries for both trained and untrained encoders? To answer such questions, we analyse perplexity and ROUGE scores of random HRED and fully trained HREDs of various hidden sizes. We go beyond the NLP classification tasks on which random embeddings were shown to be useful \cite{notraining} by testing its efficacy on a conditional language generation task.

Our main contribution is to present empirical evidence that using random projections to represent text hierarchy can achieve results on par with fully trained representations. Even without powerful pretrained word embeddings, we show that random hierarchical representations of an input text perform similarly to trained hierarchical representations. We also empirically demonstrate that, for general seq2seq models with attention, the gap between random encoder and trained encoder becomes smaller with increasing size of representations. We finally provide an evidence to validate that optimization and training of our networks was done properly. To the best of our knowledge, it is the first time that such analysis has been performed on a general class of seq2seq with attention and for the challenging task of long text summarization.

\section{Related Work}

\subsection{Fixed random weights in neural networks}

A random neural network \cite{Minsky1961LearningIR} can be defined as a neural network whose weights are initialized randomly or pseudo-randomly and are not trained or optimized for a particular task. Random neural networks have been studied since training and optimization procedures were often infeasible with the computational resources at the time. It was shown that, for low dimensional problems, Feed Forward Neural Networks (FFNN) with fixed random weights can achieve comparable accuracy and smaller standard deviations compared to the same network trained with gradient backpropagation \cite{201708}. Inspired by this work, Extreme Learning Machines (ELM) have been proposed \cite{1380068}. ELM is a single layer FFNN where only the output weights are learned through simple generalized inverse operations of the hidden layer output matrices. Subsequent theoretical studies have demonstrated that even with randomly generated hidden weights, ELM maintains the universal approximation capability of the equivalent fully trained FFNN \cite{Huang:2006:UAU:2325820.2327653}. Such works explored the effects of randomness in vision tasks with stationary models. In our work, we explore randomness in NLP tasks with autoregressive models.

Similar ideas have been developed for RNNs with Echo State Networks (ESN) \cite{esn_article} and more generally Reservoir Computing (RC) \cite{Krylov_2018}. At RC's core, the dynamics of an input sequence are modeled by a large reservoir with random, untrained weights whose state is mapped to an output space by a trainable readout layer. ESN leverage the Markovian architectural bias of RNNs and are thus able to encode input history using recurrent random projections. ESN has comparable generalization to ELM and is generally known to be more robust for non-linear time series prediction problems \cite{comp_esn_elm}. In such research, randomness was used in autoregressive models but not within the context of encoder-decoder architectures in NLP.

\begin{figure*}[t]
    \center{\includegraphics[scale=0.2]
        {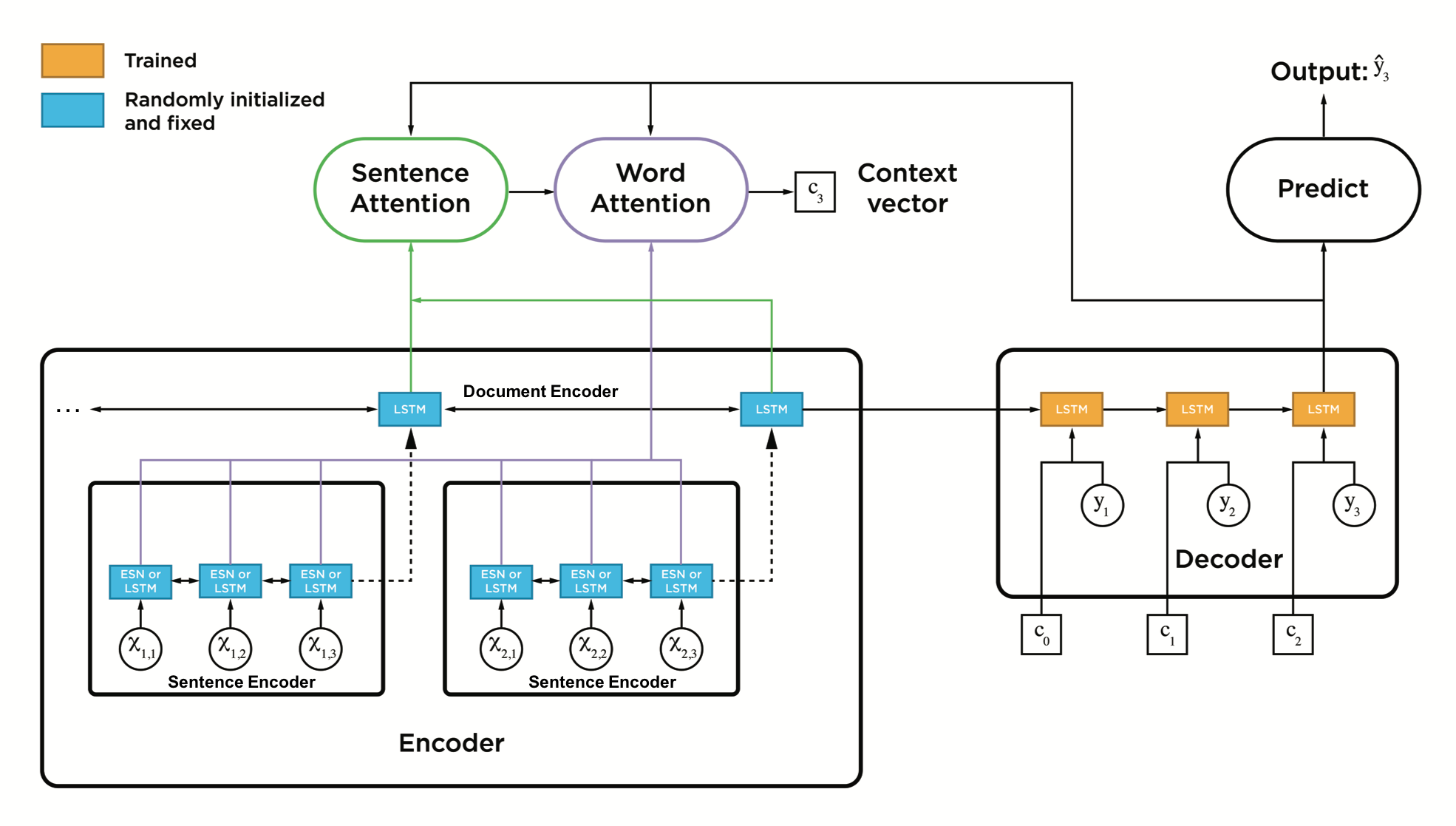}}
        \caption{The architecture of a random encoder summarization model. The weight parameters of sentence and document encoder LSTMs are randomly initialized and fixed. Other parameters for word embeddings, word and sentence level attention, and decoder LSTMs are learned during training. The blue parts of the architecture are encoder recurrent neural networks whose weights have been randomly initialized and that are not trained. The orange parts are decoder LSTMs whose weights are trained. \(\mathrm{c}\) is the encoder context vector, \(\textrm{y}\) is the ground truth target summary token, and \(\hat{\textrm{y}}\) is the predicted token.}
        \label{fig_model}
\end{figure*}

\subsection{Random encoders in deep architectures}
\label{sect:rand_enc}
Fixed random weights have been studied to encode various types of input data. In computer vision, it was shown that random kernels in Convolutional Neural Networks (CNN) perform reasonably well on object recognition tasks \cite{a2547565eaa84fefb008ae6383345dec}. Other works highlighted the importance of setting random weights in CNN and found that performance of a network could be explained mostly by the choice of a CNN architecture, instead of its optimized weights \cite{Saxe:2011:RWU:3104482.3104619}. 

Similarly, random encoder architectures in Natural Language Processing (NLP) were deeply investigated in \cite{notraining} in the context of sentence embeddings. In their experiments, pre-trained word embeddings were passed through 3 different randomly weighted and untrained encoders: a bag of random embedding projections, a random Long Short Term Memory Network \cite{Hochreiter:1997:LSM:1246443.1246450} (rand-LSTM) and an echo state network (ESN). The random sentence representation was passed through a learnable decoder to solve SentEval \cite{conneau2018senteval} downstream tasks (sentiment analysis, question-type, product reviews, subjectivity, opinion polarity, paraphrasing, entailment and semantic relatedness). Interestingly, the performance of random sentence representations was close to other modern sentence embeddings such as InferSent \cite{conneau2017supervised} and Skipthought \cite{kiros2015skip}. The authors argued that the effectiveness of current sentence embedding methods seem to benefit largely from the representational power of pre-trained word embeddings. While random encoding of the input text was deployed to solve NLP classification tasks, random encoding for conditional text generation has still not been studied.

\section{Approach}
We hypothesize that the Markovian representation of word history and position within the text, as provided by randomly parameterized encoders, is rich enough to achieve comparable results to fully trained networks, even for difficult NLP tasks. In this work, we compare Hierarchical Recurrent Encoder Decoder (HRED) models with randomly fixed encoder with ones fully trained in a normal end-to-end manner on abstractive summarization tasks. In particular, we examine the performances of random encoder models with varying (1) encoder and (2) decoder capacity. To isolate root causes of performance gaps between trained encoders and random untrained encoders, we also provide an analysis of gradient flows and relative weight changes during training.

\begin{table*}[t]
\begin{center}
\begin{tabular}{|c|c|ccc|c}
	\hline 
		\multirow{2}*{Model} & \multirow{2}*{Encoder} 
		&  \multicolumn{3}{c|}{ROUGE}  \\
        & & 1 & 2 & L  \\ \hline
        abstractive model \cite{nallapati2016abstractive} & Trained Hierarchical GRU & 35.46 & 13.30 & 32.65 \\
        seq2seq + attn (150K vocab)  & Trained LSTM & 30.49 & 11.17 & 28.08 \\   
        seq2seq + attn (50K vocab) & Trained LSTM & 31.33 & 11.81 & 28.83  \\
        pointer-generator network \cite{asee} & Trained LSTM & 36.44 & 15.66 & 33.42 \\ \hline\hline
        HRED + attn + pointer (ours) & Trained H-LSTM & 35.72 & 15.08 & 32.85 \\
        HRED + attn + pointer (ours) & Random H-LSTM & 34.51 & 13.89 & 31.59 \\
        HRED + attn + pointer (ours) & Random LSTM + ESN & 34.60 & 13.98 & 31.74 \\
        HRED + attn + pointer (ours) & Identity H-LSTM & 27.11 & 8.14 & 25.16 \\
    \hline
\end{tabular}
\end{center}
\caption{Results of the CNN / Daily Mail test dataset. ROUGE scores have a 95\% confidence interval of at most \textpm0.24 as reported by the official ROUGE script.}
\label{tab:rouge_compare}
\end{table*}

\subsection{Model}
\label{sect:hred}
Our hierarchical recurrent encoder decoder model is similar to that of \cite{nallapati2016abstractive}. The model consists of two encoders: sentence encoder and document encoder. The sentence encoder is a recurrent neural network which encodes a sequence of input words in a sentence into a fixed size sentence representation. Specifically, we take the last hidden state of the recurrent neural network as the sentence encoding.
\begin{equation} \label{eq:rnn_sen}
    h^{(s)} = \textrm{RNN}_{sen}(x_{1}, x_{2}, \ldots, x_{N})
\end{equation}
where \(x_{i}\) are embeddings of \(i\) th input token and \(N\) is the length of the corresponding sentence. The sentence encoder is shared for all sentences in the input document. The sequence of sentence encodings are then passed to the document encoder which is another recurrent neural network.
\begin{equation}
    h^{(d)} = \textrm{RNN}_{doc}(h^{(s_{1})}, h^{(s_{2})}, \ldots, h^{(s_{M})})
\end{equation}
where \(h^{(s_j)}\) denotes the encoding of \(j\) th sentence and \(M\) is the number of sentences in the input document.

The decoder is another recurrent neural network that generates a target summary token by token. To capture relevant context from the source document, the decoder leverages the same hierarchical attention mechanism used in \cite{nallapati2016abstractive}. Concretely, the decoder computes word-level attention weight \(\beta\) by using the sentence encoder states of input tokens. The decoder also obtains sentence-level attention weight \(\gamma\) using document encoder hidden states. The final attention weight \(\alpha\) integrates word- and sentence-level attention to capture salient part of the input in both word and sentence levels.
\begin{equation}
    \alpha_{k} = \frac{\beta_{k}\gamma_{s(k)}}{\sum_{l=1}^{N_{d}} \beta_{l}\gamma_{s(l)}}
\end{equation}
where \(\beta_{k}\) and \(\beta_{l}\) denote the word-level attention weight on the \(k\) th and \(l\) th tokens of the input document respectively, \(\gamma_{m}\) is the sentence-level attention weight for the \(m\) th sentence of the input document, \(s(l)\) returns the index of the sentence at the \(l\) th word position and \(N_d\) is the total number of tokens in the input document.

A pointer-generator architecture enables our decoder to copy words directly from the source document. The use of pointer-generator allows the model to effectively deal with out-of-vocabulary tokens. Additionally, decoder coverage is used to prevent the summarization model from generating the same phrase multiple times. The detailed description of pointer-generator and decoder coverage can be found in \cite{arxivpubmed}.

\section{Experiment and Analysis}
In our experiments, we aim to demonstrate that random encoding can reach similar performances of trained encoding in a conditional natural language generation task. To appreciate this contribution, we will first describe the experimental and architectural setup before deep diving into the results. The CNN/Daily Mail dataset\footnote{We used the data and preprocessing code provided in https://github.com/abisee/cnn-dailymail}  \cite{Hermann:2015:TMR:2969239.2969428,nallapati2016abstractive} is used for the summarization task. For fully-trainable hierarchical encoder models, we use two bi-directional LSTMs for the sentence and the document encoder (Trained H-LSTM). There are 3 different types of untrained, random hierarchical encoders that we investigate: 

\begin{enumerate}[label=(\alph*)]
\item Random H-LSTM: Random bi-directional LSTMs (rand-LSTM) for the sentence and document encoder, with weight matrices and biases initialized uniformly from $U(-\frac{1}{\sqrt{d}}, \frac{1}{\sqrt{d}})$ where $d$ is the hidden size of LSTMs.
\item Identity H-LSTM: Similar to (a) but with LSTM hidden weights and biases matrices set to the identity $\mathcal{I}$.
\item Random LSTM + ESN: A random bi-directional LSTM sentence encoder initialized in the same way as (a) and an echo state network (ESN) for the document encoder, with weights sampled i.i.d. randomly from the Normal distribution $N(0,1)$.
\end{enumerate}

The architecture of a random encoder summarization model is depicted in Figure~\ref{fig_model}. All recurrent networks including the echo state network have a single layer. Note that by using tied embeddings \cite{press2016tied}, source word embeddings are learned in both random and trainable encoders. This is an important setting in our experiments as we aim to isolate the effect of trained encoders from that of trained word embeddings. In all experiments, a single-directional LSTM is used for the decoder. We generally follow the standard hyperparameters suggested in \cite{asee} and \cite{arxivpubmed}. We guide the readers to the appendix for more details on training and evaluation steps.

\begin{table}[t] 
\begin{center}
\begin{tabular}{|c|c|ccc|}
	\hline 
		\multirow{2}*{\small{Enc}} & \multirow{2}*{\small{Dec}} & \small{Trained} & \small{Random} & \small{Random LSTM}  \\
        & & \small{H-LSTM} & \small{H-LSTM} & \small{+ ESN}  \\ \hline
        \multirow{3}*{\scriptsize{256}} & \scriptsize{64} & \scriptsize{17.07} & \scriptsize{22.17} \scriptsize{(+30\%)} & \scriptsize{21.84} \scriptsize{(+28\%)} \\
        & \scriptsize{256} & \scriptsize{14.81} & \scriptsize{18.03} \scriptsize{(+22\%)} & \scriptsize{18.04} \scriptsize{(+22\%)} \\
        & \scriptsize{1024} & \scriptsize{14.47} & \scriptsize{16.75} \scriptsize{(+19\%)} & \scriptsize{16.84} \scriptsize{(+20\%)} \\ 
        \hline\hline
        \scriptsize{64} & \multirow{3}*{\scriptsize{256}} & \scriptsize{15.66} & \scriptsize{22.75} \scriptsize{(+45\%)} & \scriptsize{22.43} \scriptsize{(+43\%)} \\
        \scriptsize{256} &  & \scriptsize{14.81} & \scriptsize{18.03} \scriptsize{(+22\%)} & \scriptsize{18.04} \scriptsize{(+22\%)} \\
        \scriptsize{1024} &  & \scriptsize{14.02} & \scriptsize{17.09} \scriptsize{(+18\%)} & \scriptsize{17.16} \scriptsize{(+19\%)} \\
        \hline
\end{tabular}
\end{center}
\caption{Test perplexity of trained and random hierarchical encoder models on the CNN / Daily Mail dataset (lower is better). Note that Enc = encoder hidden size and Dec = decoder hidden size. Percentages in parentheses are the relative perplexity degradation in random encoder models with respect to the associated Trained H-LSTM encoder model.}
\label{tab:perplexity_compare}
\end{table}

\subsection{Performance of random encoder models}
Table~\ref{tab:rouge_compare} shows ROUGE scores \cite{rouge}, a commonly used performance metric in summarization, for trained and untrained-random encoder models. Of note, our hierarchical random encoder models (Random H-LSTM and Random LSTM + ESN) obtain ROUGE scores close to other trained models. The gap between our trained and random hierarchical encoders is about 1.1 point for all ROUGE scores. Hierarchical random encoders even outperform a competitive baseline \cite{nallapati2016abstractive} in terms of ROUGE-2, even though the cited model uses pre-trained embeddings. With respect to the trained H-LSTM, the Random H-LSTM achieves ROUGE scores that are very similar: the gap is 3.5\% in ROUGE-1, 8.5\% in ROUGE-2 and 3.9\% in ROUGE-L. We also tested the Identity H-LSTM to get an idea of the role that trained word embeddings play in the performance. The Identity H-LSTM creates sentences representations by accumulating trained word embeddings in equation (\ref{eq:rnn_sen}). To measure the representational power of random projection in the encoder, we compare the ROUGE scores of random hierarchical encoder models with those of the Identity H-LSTM model. We notice that the random encoders greatly outperform\footnote{The performance gap between random and identity initialization could be more pronounced if the input word embeddings had not been trained.} the Identity H-LSTM encoder. This has brought us closer to gauging the effectiveness of randomly projected recurrent encoder hidden states over an accumulation of word embeddings.

\begin{figure*}
\centering
\subfloat[Document Encoder]{
    \includegraphics[width=0.47\linewidth]{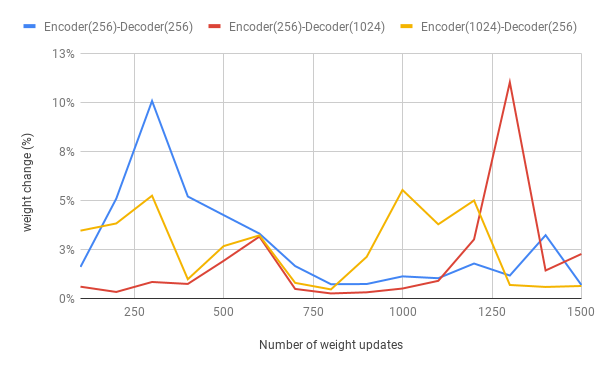}
}
\subfloat[Sentence Encoder]{
    \includegraphics[width=0.5\linewidth]{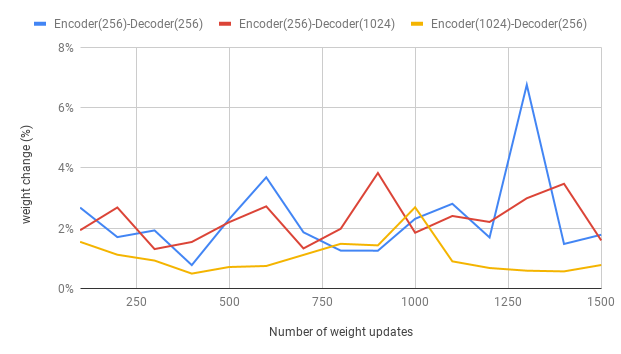}
}
\caption{Relative weight change $\big\{\frac{\Delta{w}}{w}\big\}$ at every 100 update for the encoders of the trained H-LSTM. The figure legend indicates different combinations of encoder-decoder hidden sizes.}
\label{fig:weight_change}
\end{figure*}

\subsection{Impact of increasing capacity}
Table~\ref{tab:perplexity_compare} shows the test perplexity of random hierarchical encoder models with different encoder or decoder hidden sizes. We chose to base our analysis on perplexity instead of ROUGE to isolate model performance from the effect of word sampling methods such as beam search and to show the quality of overall predicted word probabilities. It is shown that increased model capacity leads to lower test perplexity, which implies better target word prediction. The improvement in performance, however, is not equal across models. We notice that random encoders close the performance gap with the fully trained counterpart as the encoder hidden size increases. For instance, as we vary the encoder hidden size of the Random H-LSTM from 64 to 1024, the relative perplexity gap with the Trained H-LSTM diminishes from 45\% to 18\%. This pattern aligns with the previous work from \citet{notraining}, where authors discovered that the performance of random sentence encoder converged to that of trained one as the dimensionality of sentence embeddings increased. 

We perform similar experiments with the decoder hidden size which varies from 64 to 1024, while fixing the encoder hidden size to 256. We first expected that the hidden size of a trained decoder would play a larger role in enhancing the model performance than that of a random encoder. As shown in Table~\ref{tab:perplexity_compare}, however, the perplexity of random encoder models with the largest encoder hidden size (16.75 and 16.84) is close to that of random encoder models with the largest decoder hidden size (17.09 and 17.16). Moreover, the performance gaps of the previously mentioned configuration with respect to its fully trained counterpart are just as small (19\% vs 18\% and 20\% vs 19\%). There are two conclusions that we can draw from this result. First, increasing the capacity of random encoder closes the performance gap between fully trained and random encoder models. Second, increasing the number of parameters in a random encoder yields similar improvements to increasing the number of parameters of a trainable decoder. This illustrates important advantages of using random encoder in terms of the number of parameters to train.

\subsection{Gradient flow of trainable encoders}
One may suspect that the smaller performance gap in bigger encoder or decoder hidden size might arise from optimization issues in RNNs \cite{Bengio:1994:LLD:2325857.2328340, chapter-gradient-flow-2001}, such as the vanishing gradient problem. To verify that the parameters of large trained models are learned properly, we analyze the distribution of weight parameters and gradients of each model.

From our results, we first notice that networks with different capacity have different scale of parameter and gradient values. This makes it infeasible to directly compare the gradient distributions of models with different capacity. We thus examine the relative amount of weight changes in trained encoder LSTMs as follows:
\begin{equation} 
\Big\{\frac{\Delta{w}}{w}\Big\}_{i} = \frac{1}{N} \sum_{j=0}^{N}{\abs{\frac{w_{j, i+100} - w_{j, i}}{w_{j, i}}}}
\end{equation}
where \(w\) is a weight parameter of the encoder LSTM, $N$ is the number of parameters in the encoder LSTM, $j$ is the weight index and $i$ is the number of training updates. The relative encoder's weight change is depicted in Figure~\ref{fig:weight_change} over 1500 iterations. We observe that there is no significant difference in the relative weight changes between small and large encoder models. Sentence and document encoder weights with 256 and 1024 hidden sizes show similar patterns over training iterations. For more details on the distributions of weight parameters and gradients, we refer the reader to the appendix. We have also added training curves in the appendix to show that our trained models indeed converged. Given that the parameters of the trained H-LSTM were properly optimized, we can thus conclude that the trained weights do not contribute significantly to model performance on a conditional natural language generation task such as summarization.

\section{Conclusion and future work}
In this comparative study, we analyzed the performance of untrained randomly initialized encoders on a more complex task than classification \cite{notraining}. Concretely, the performance of random hierarchical encoder models was evaluated on the challenging task of abstractive summarization. We have shown that untrained, random encoders are able to capture the hierarchical structure of documents and that their summarization qualiy is comparable to that of fully trained models. We further provided empirical evidence that increasing the model capacity not only enhances the performance of the model but closes the gap between random and fully trained hierarchical encoders. For future works, we will further investigate the effectiveness of random encoders in various NLP tasks such as machine translation and question answering.

\section{Acknowledgements}
We would like to thank Devon Hjelm, Ioannis Mitliagkas, Catherine Lefebvre for their useful comments and Minh Dao for helping with the figures. This work was partially supported by the IVADO Excellence Scholarship and the Canada First Research Excellence Fund.

\bibliography{acl2019}
\bibliographystyle{acl_natbib}

\cleardoublepage
\appendix

\section{Appendix}
\label{sec:supplemental}

\subsection{Training and Evaluation}
\label{sect:train_evaluation}
The dimensionality of embeddings is 128 and embeddings are trained from scratch. The vocabulary size is limited to 50,000. During training, we constrain the document length to 400 tokens and the summary length to 100 tokens. Batch size is 8 and learning rate is 0.15. Adagrad \cite{duchi2011adaptive} with an initial accumulator value of 0.1 is used for optimization. Maximum gradient norm is set to 2. Training is performed for 12 epochs. At test time, we set the maximum number of generated tokens to 120. Beam search with beam size 4 is used for decoding. To evaluate the qualities of generated summaries, we use the standard ROUGE metric \cite{rouge} and report standard F-1 ROUGE scores. 

\subsection{Learning curves}
\begin{figure*}
\centering
\subfloat[Enc(64)-Dec(256)]{
  \includegraphics[width=50mm]{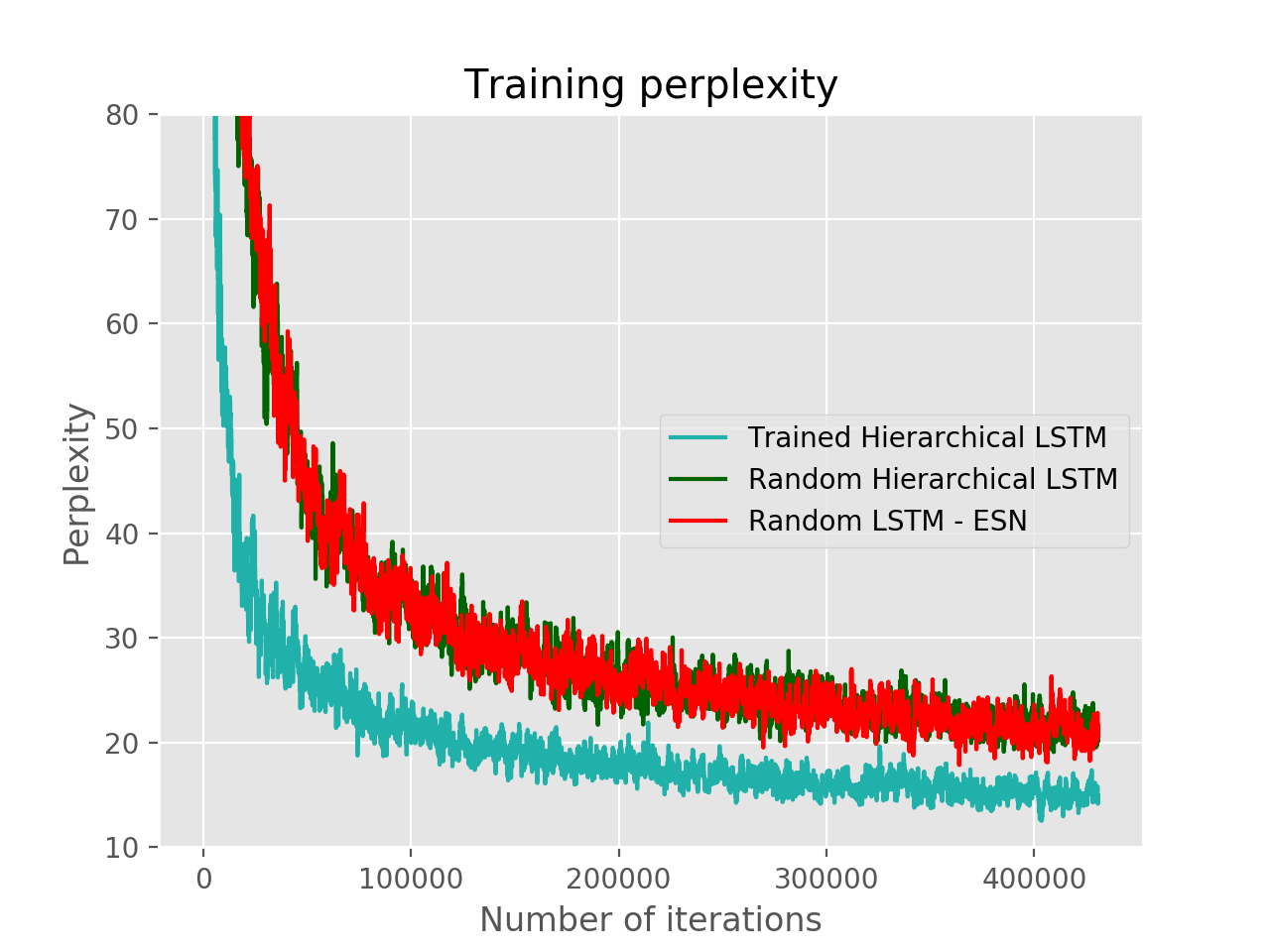}
}
\subfloat[Enc(256)-Dec(256)]{
  \includegraphics[width=50mm]{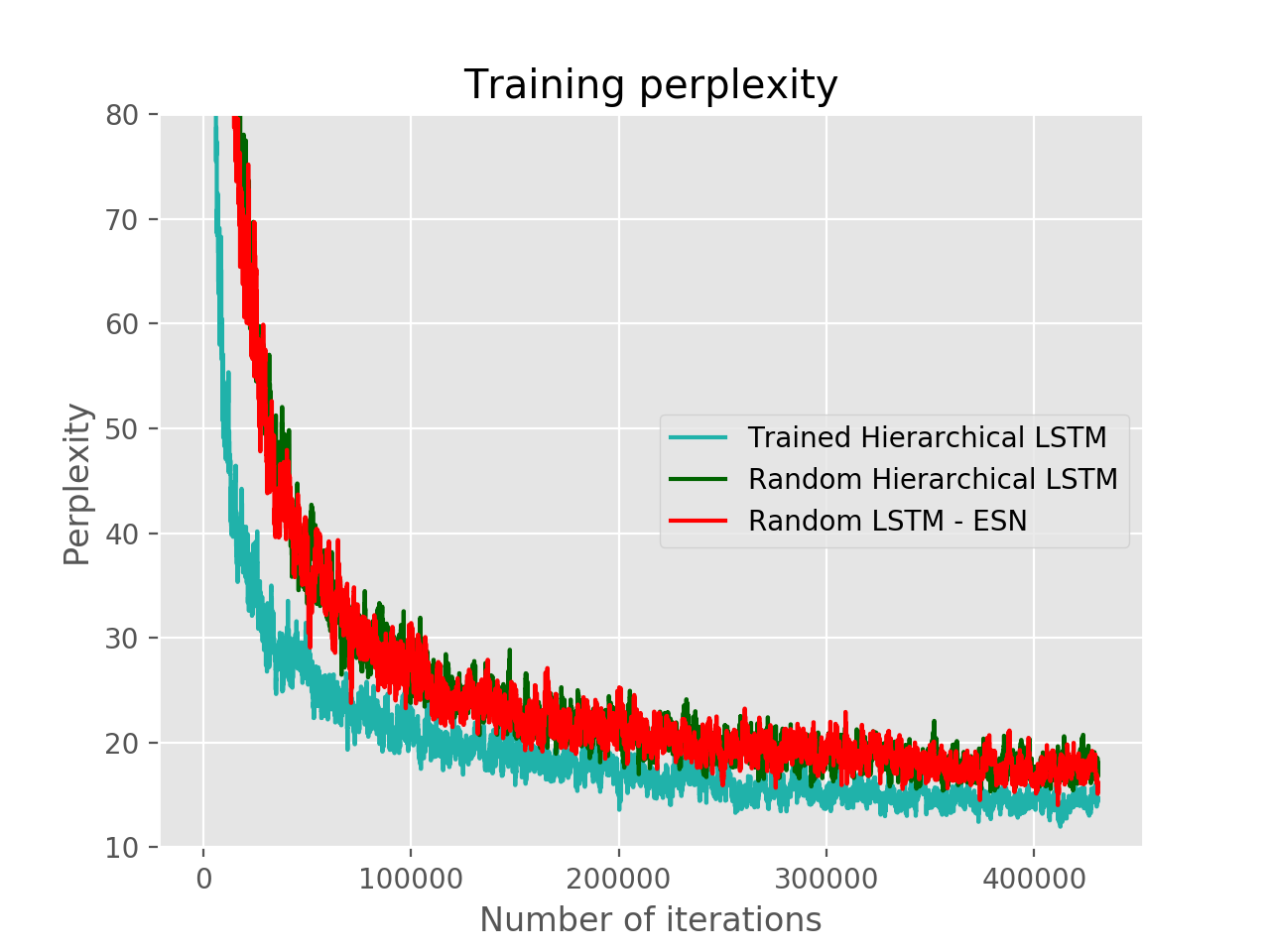}
}
\subfloat[Enc(1024)-Dec(256)]{
  \includegraphics[width=50mm]{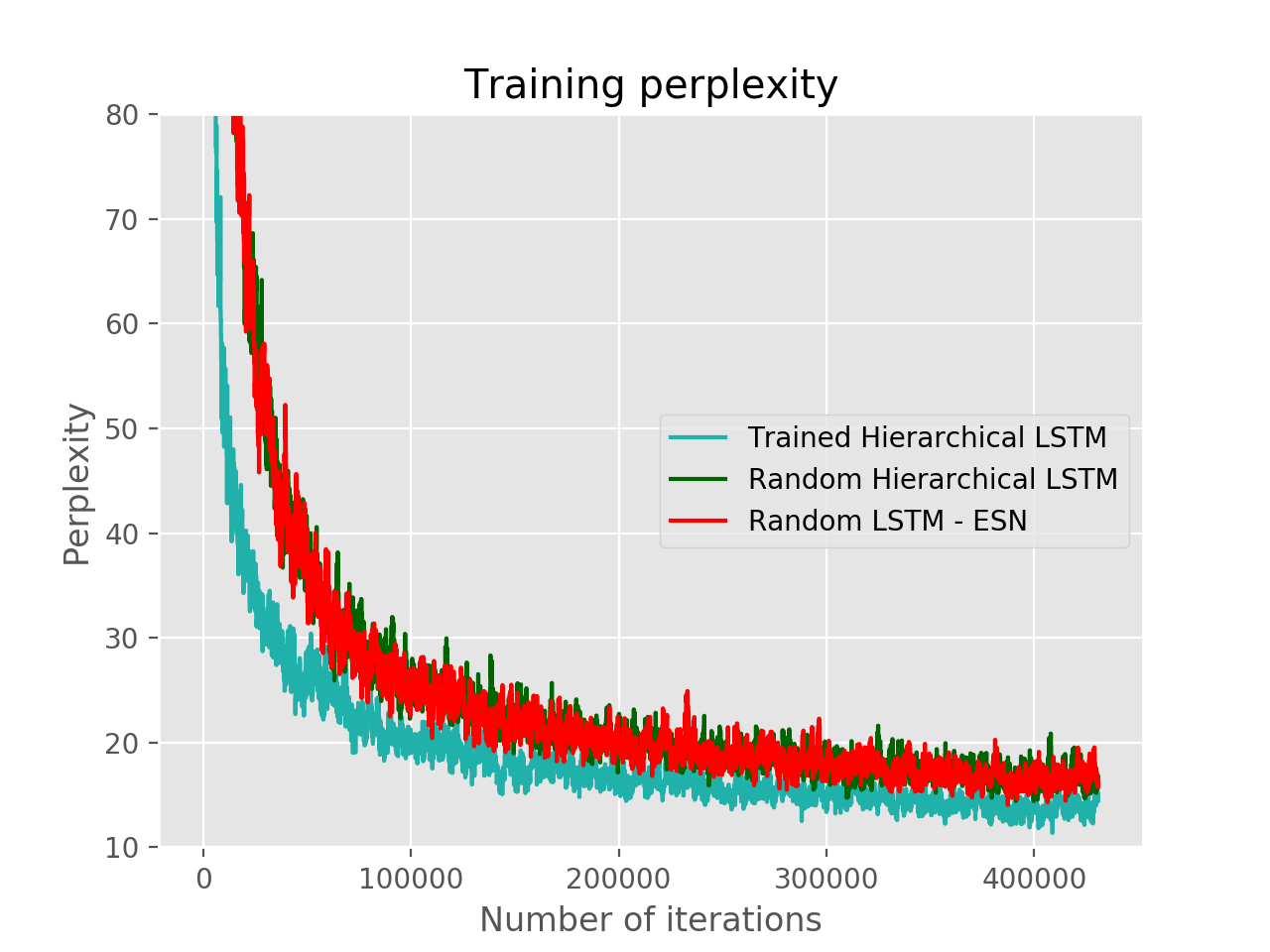}
}
\hspace{0mm}
\subfloat[Enc(256)-Dec(64)]{
  \includegraphics[width=50mm]{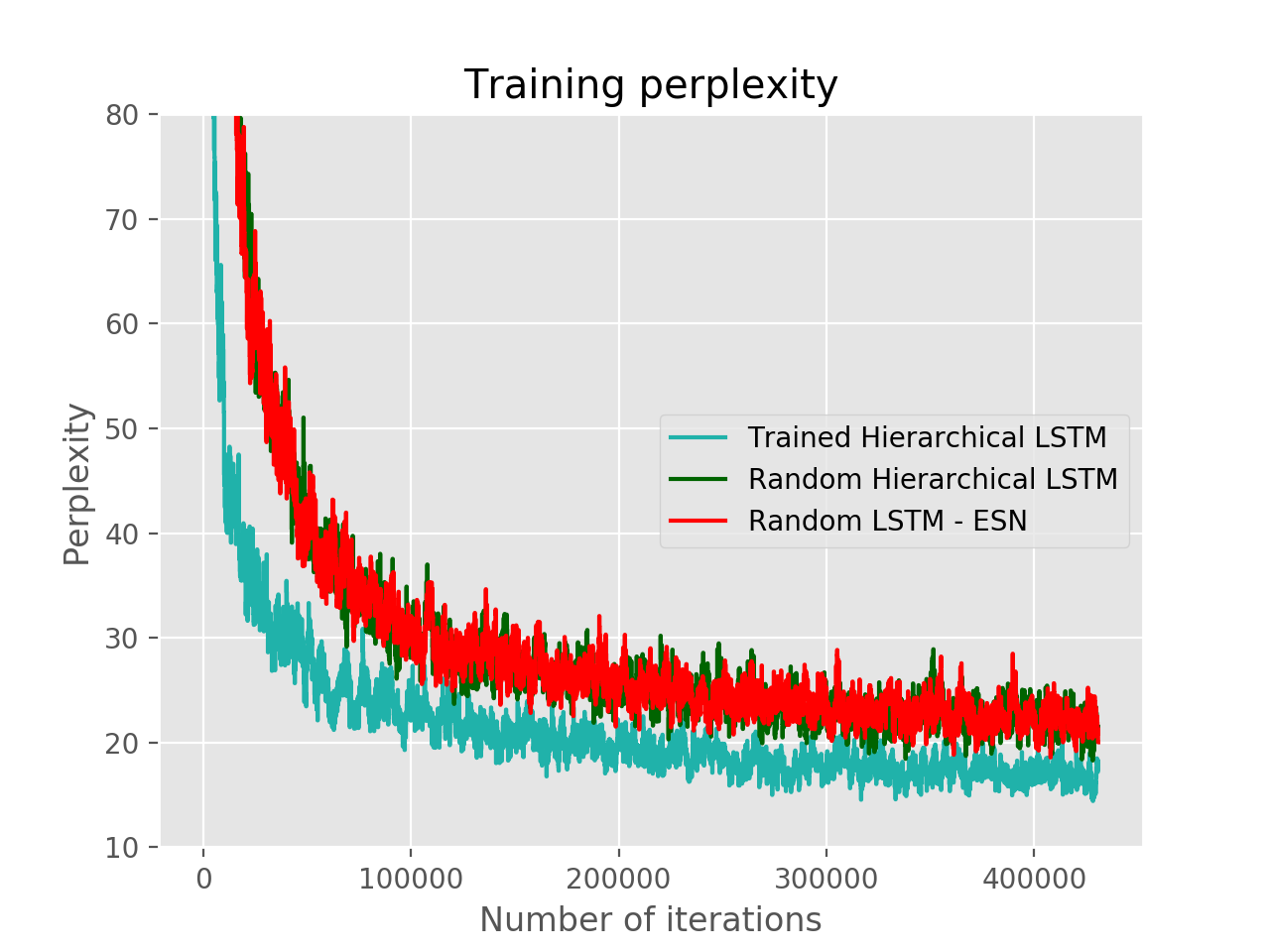}
}
\subfloat[Enc(256)-Dec(256)]{
  \includegraphics[width=50mm]{figures/curve/train_enc256_dec256.png}
}
\subfloat[Enc(256)-Dec(1024)]{
  \includegraphics[width=50mm]{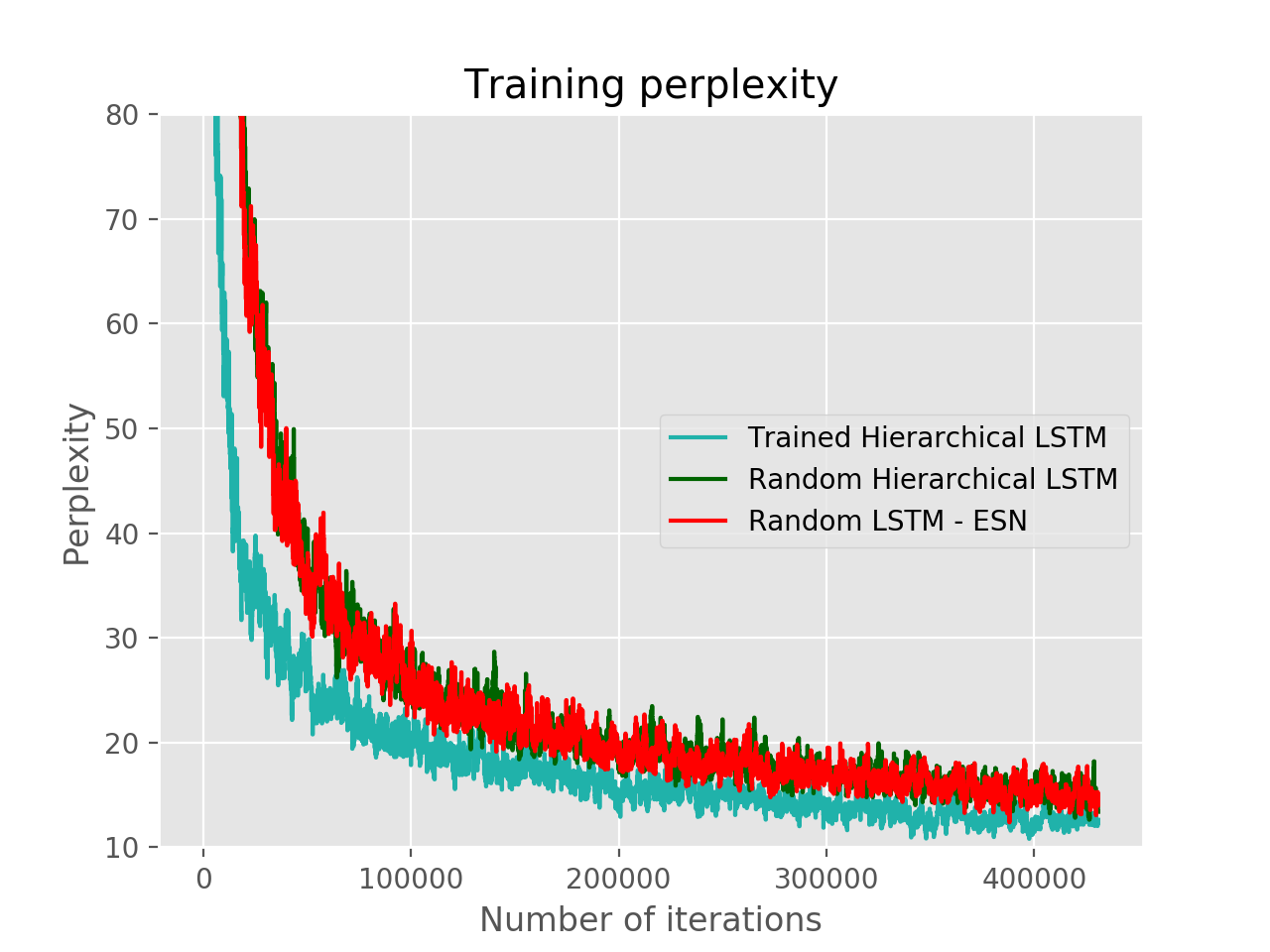}
}
\caption{Training perplexity of trained and random encoder summarization models with different encoder and decoder hidden sizes. Enc denotes encoder and Dec denotes decoder. Numbers in parentheses are corresponding hidden sizes.}
\label{fig:learning_curve_train}
\end{figure*}

\begin{figure*}
\centering
\subfloat[Enc(64)-Dec(256)]{
  \includegraphics[width=50mm]{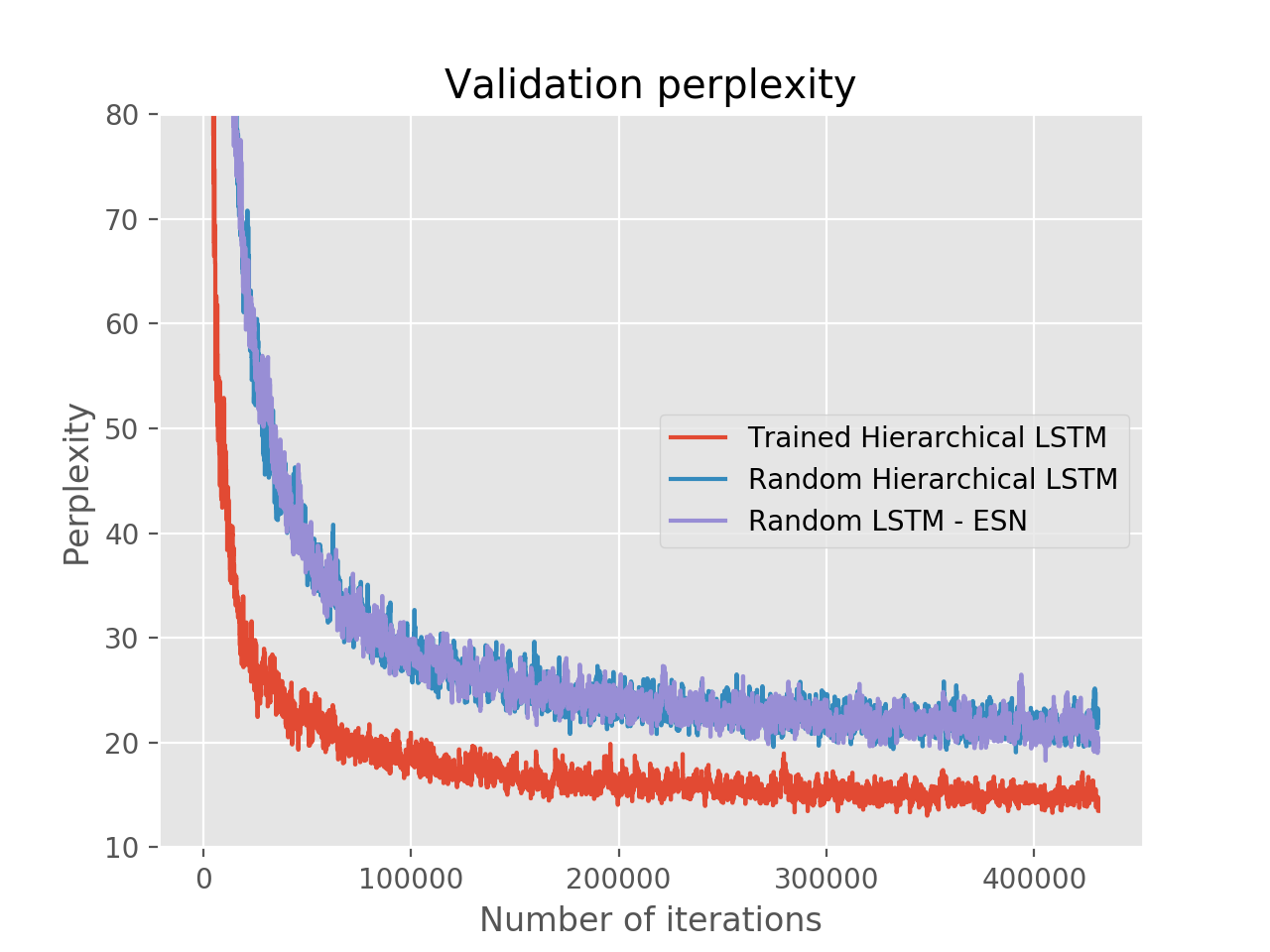}
}
\subfloat[Enc(256)-Dec(256)]{
  \includegraphics[width=50mm]{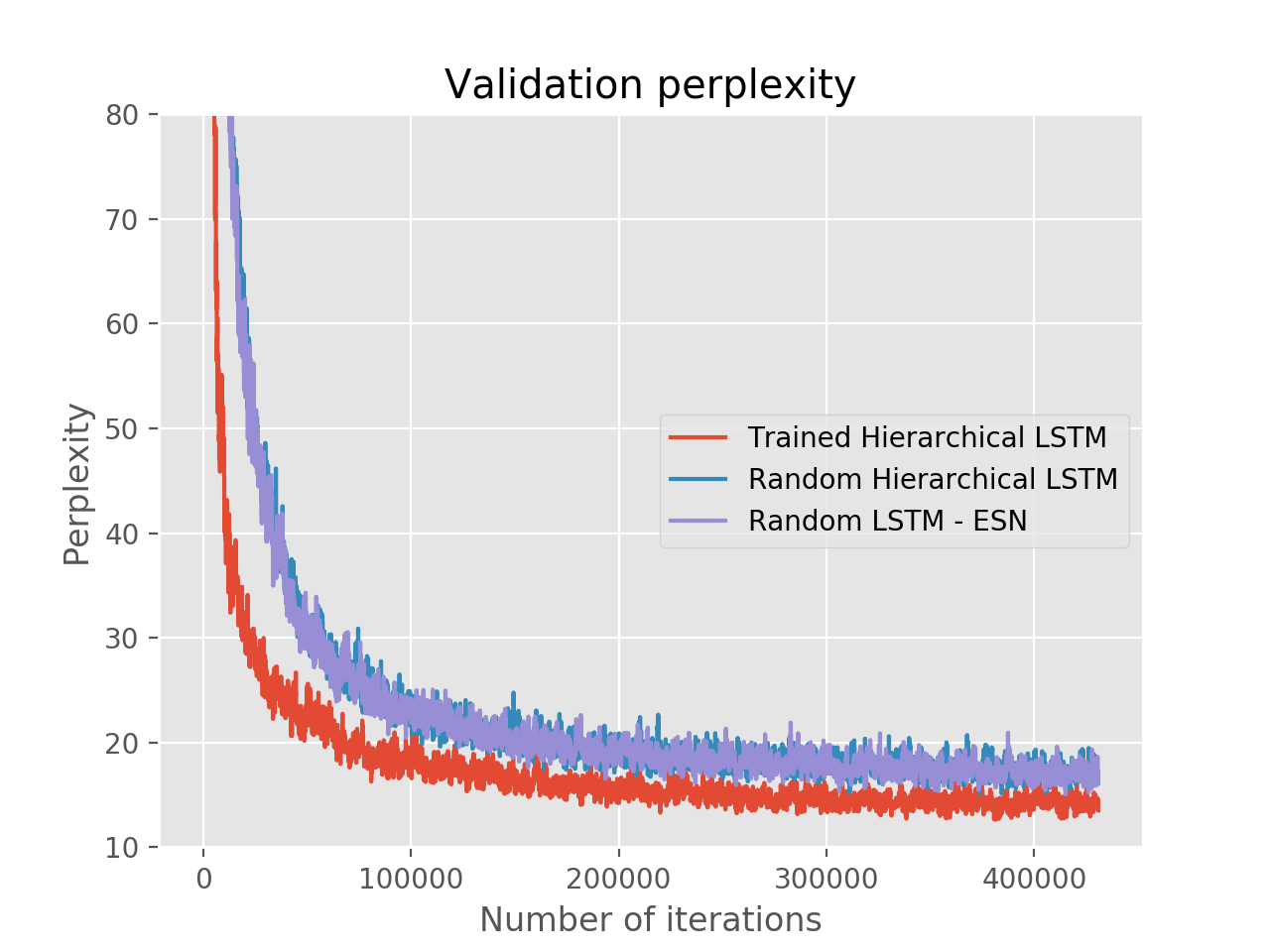}
}
\subfloat[Enc(1024)-Dec(256)]{
  \includegraphics[width=50mm]{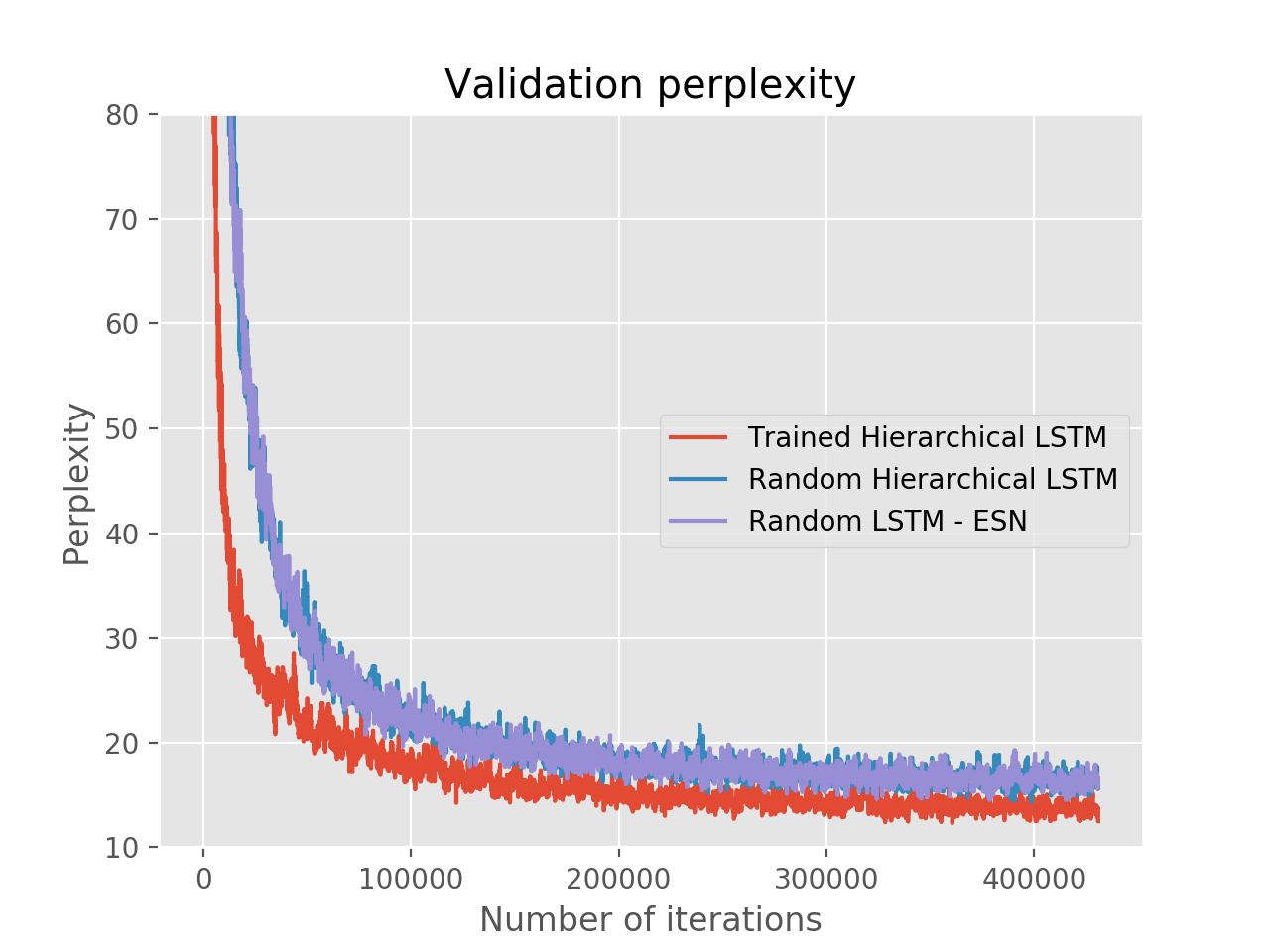}
}
\hspace{0mm}
\subfloat[Enc(256)-Dec(64)]{
  \includegraphics[width=50mm]{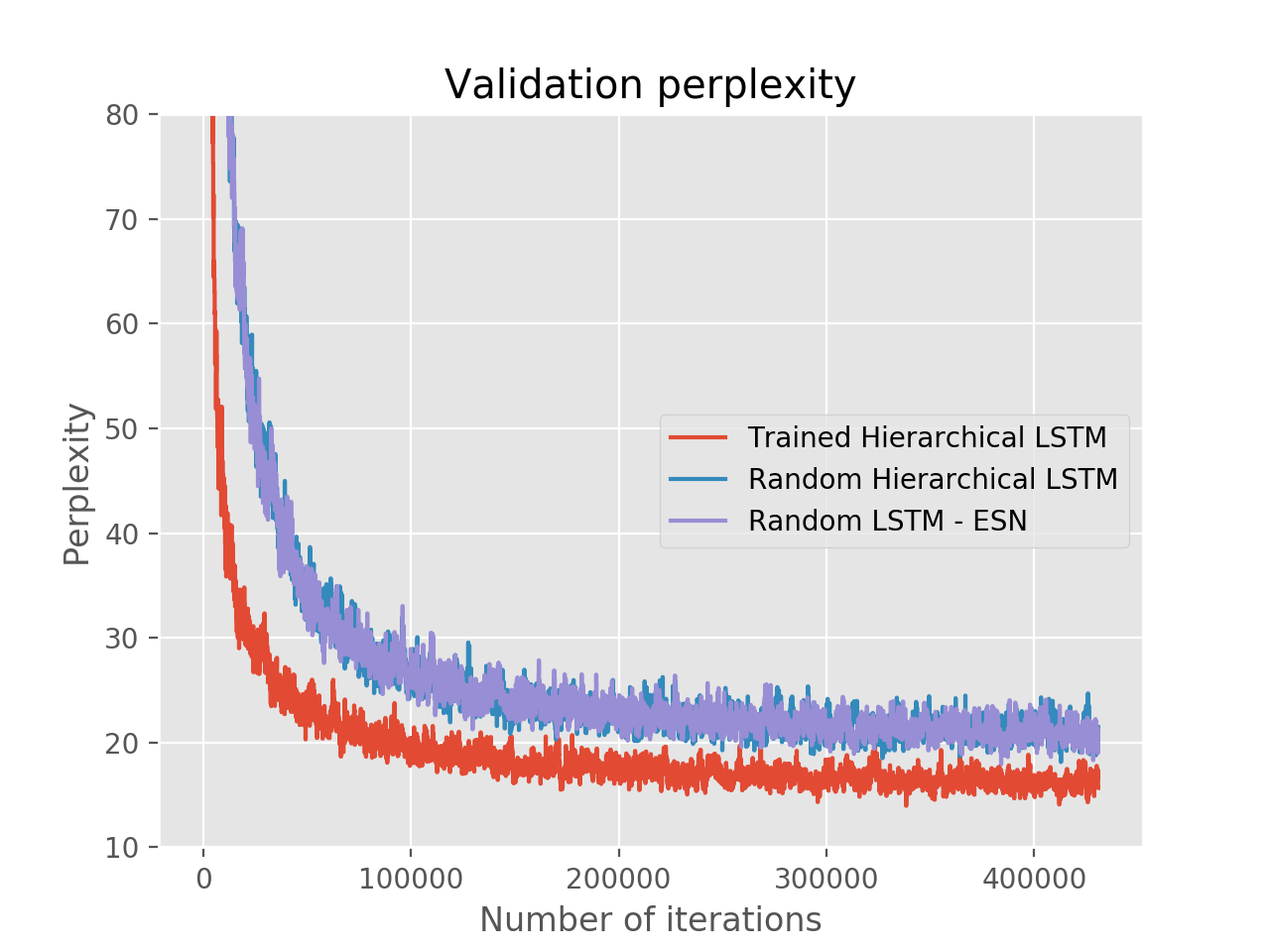}
}
\subfloat[Enc(256)-Dec(256)]{
  \includegraphics[width=50mm]{figures/curve/val_enc256_dec256.png}
}
\subfloat[Enc(256)-Dec(1024)]{
  \includegraphics[width=50mm]{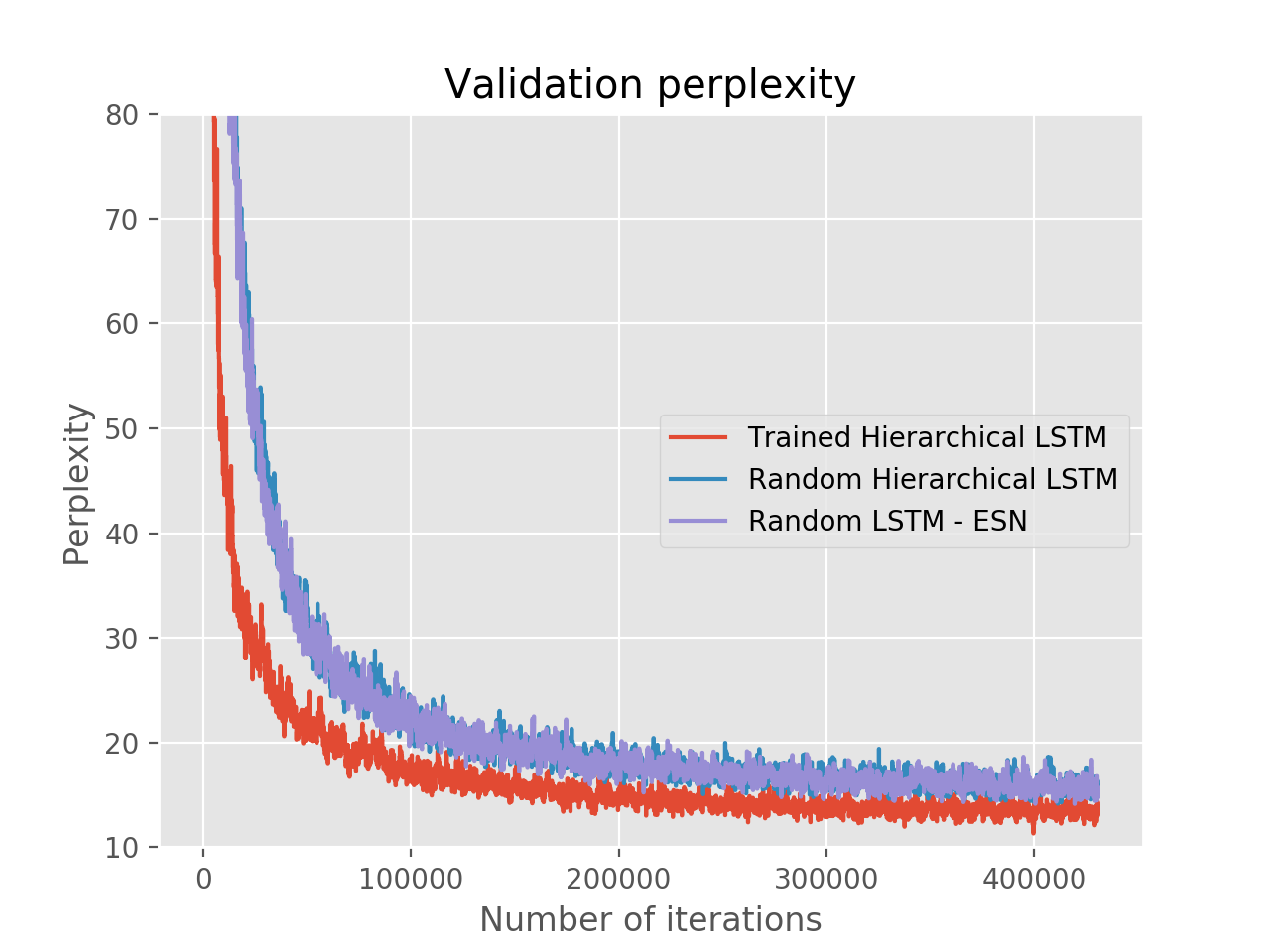}
}
\caption{Validation perplexity of trained and random encoder summarization models with different encoder and decoder hidden sizes. Enc denotes encoder and Dec denotes decoder. Numbers in parentheses are corresponding hidden sizes.}
\label{fig:learning_curve_val}
\end{figure*}

Figure~\ref{fig:learning_curve_train} and \ref{fig:learning_curve_val} show the learning curves of trainable and random encoder summarization models with different encoder and decoder hidden sizes. Note that the gap in training and validation perplexity between trained and random encoder models get smaller as the encoder or decoder hidden size increases.

\subsection{Weight and gradient distribution}
\begin{figure*}
\centering
\subfloat[Enc(64)-Dec(256) weight]{
  \includegraphics[width=45mm]{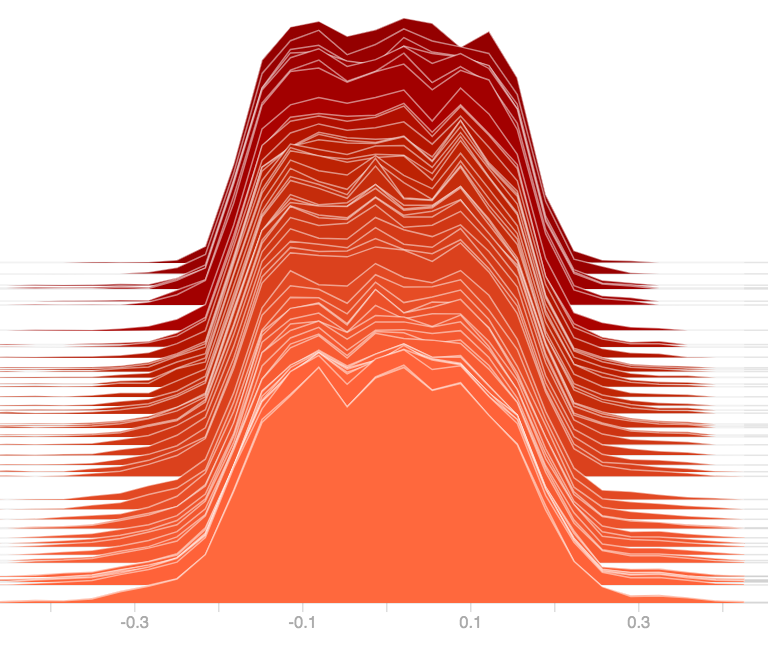}
}
\subfloat[Enc(256)-Dec(256) weight]{
  \includegraphics[width=45mm]{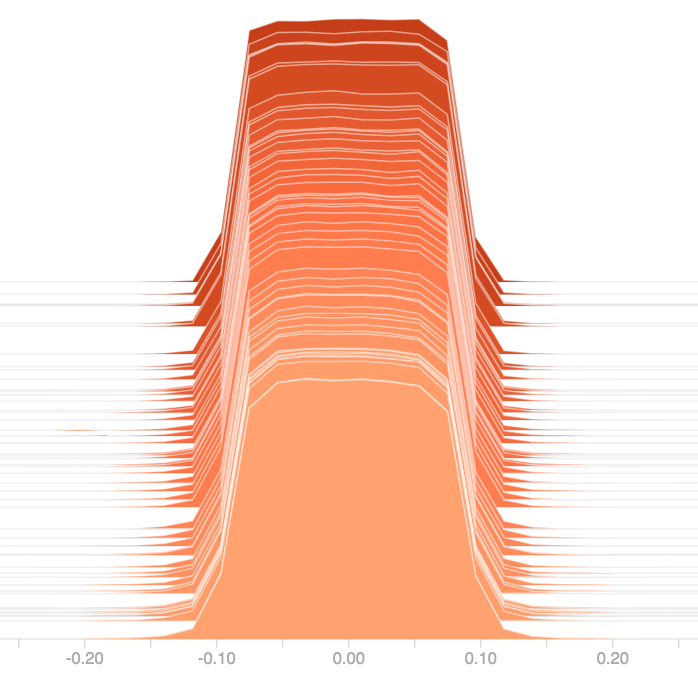}
}
\subfloat[Enc(1024)-Dec(256) weight]{
  \includegraphics[width=45mm]{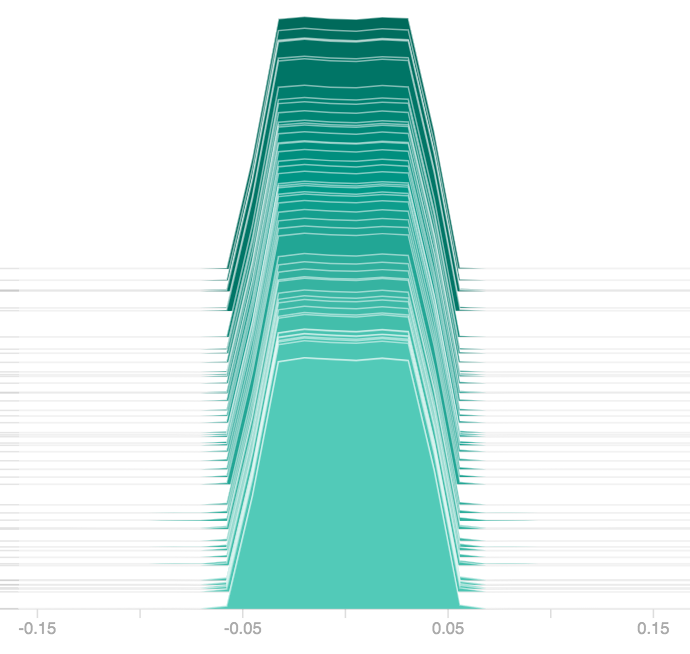}
}
\hspace{0mm}
\subfloat[Enc(64)-Dec(256) gradient]{
  \includegraphics[width=45mm]{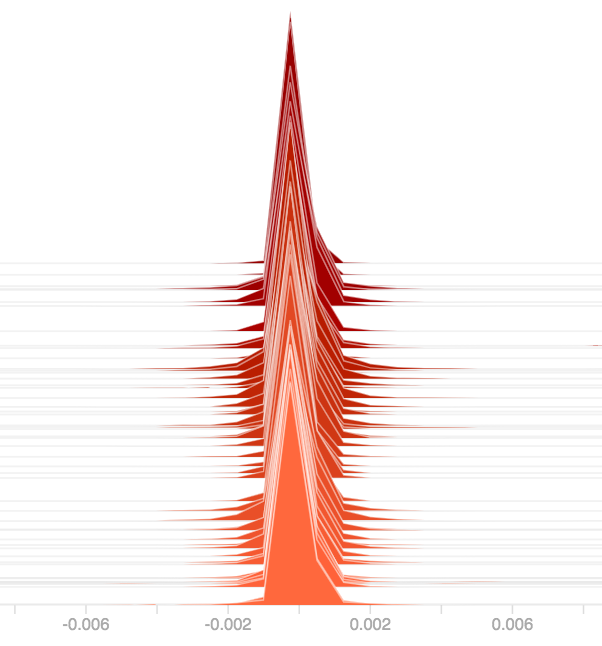}
}
\subfloat[Enc(256)-Dec(256) gradient]{
  \includegraphics[width=45mm]{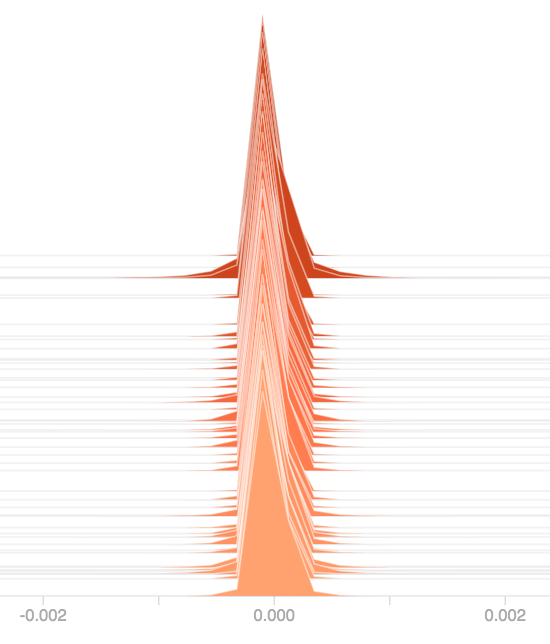}
}
\subfloat[Enc(1024)-Dec(256) gradient]{
  \includegraphics[width=45mm]{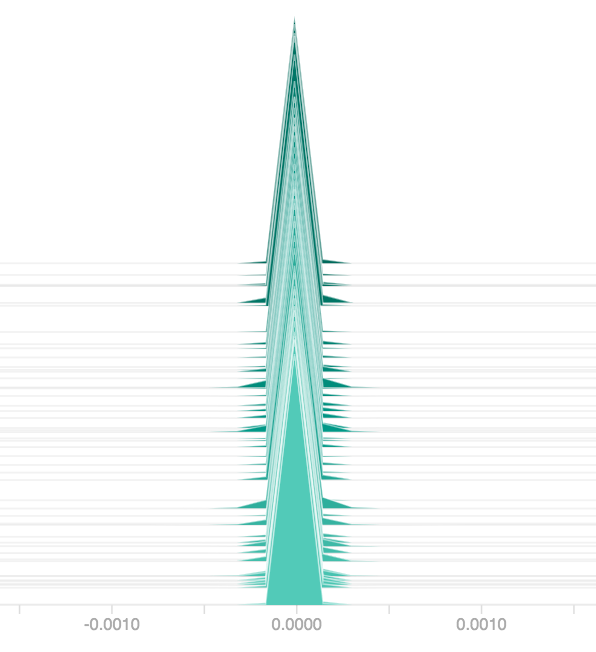}
}
\caption{Distribution of weight parameters and gradients of the document encoder.}
\label{fig:docenc_dist}
\end{figure*}

\begin{figure*}
\centering
\subfloat[Enc(64)-Dec(256) weight]{
  \includegraphics[width=45mm]{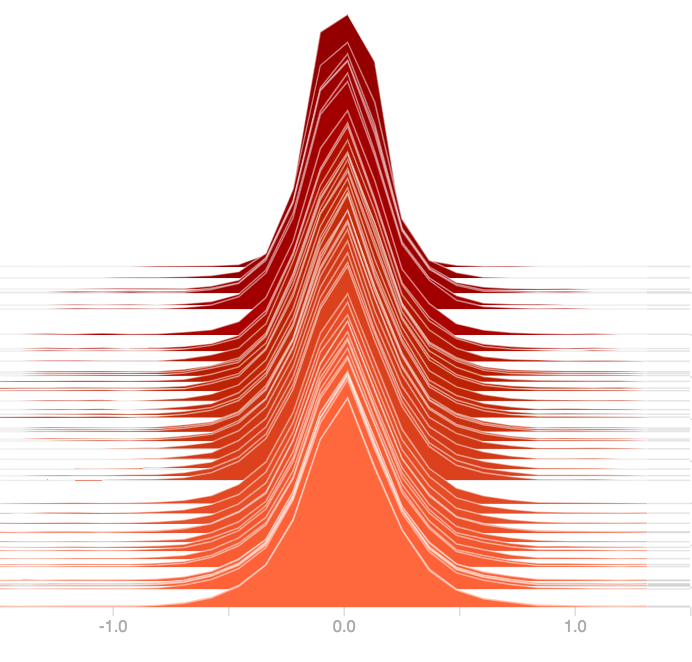}
}
\subfloat[Enc(256)-Dec(256) weight]{
  \includegraphics[width=45mm]{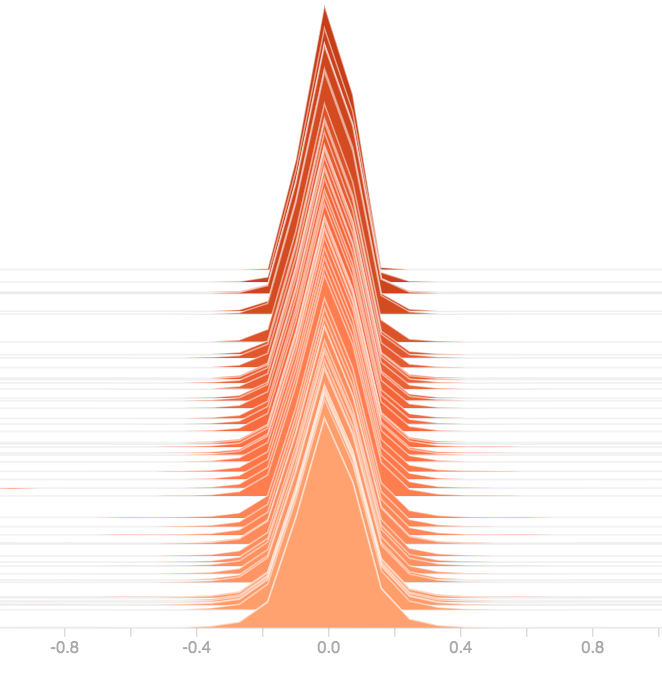}
}
\subfloat[Enc(1024)-Dec(256) weight]{
  \includegraphics[width=45mm]{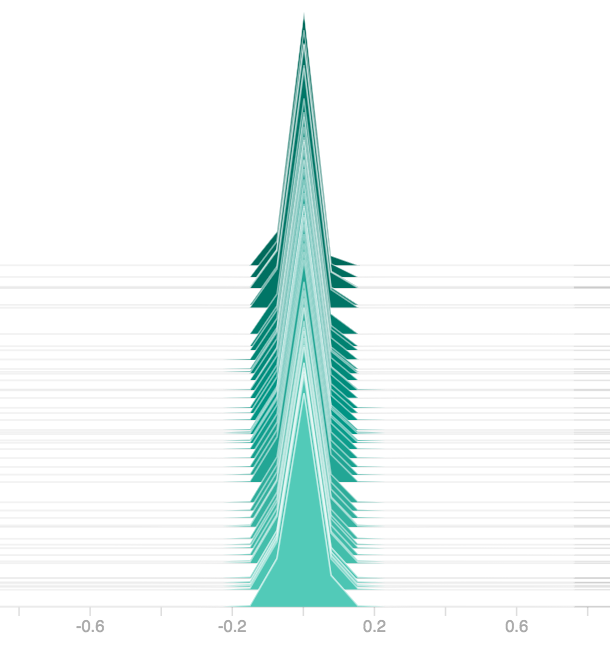}
}
\hspace{0mm}
\subfloat[Enc(64)-Dec(256) gradient]{
  \includegraphics[width=45mm]{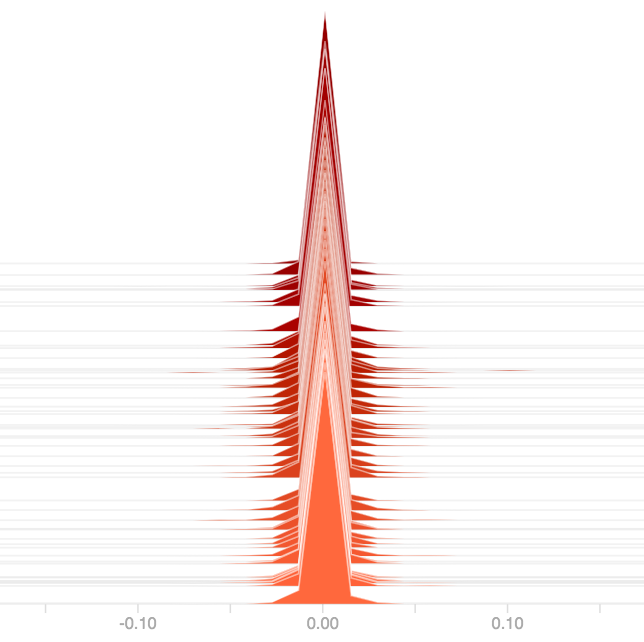}
}
\subfloat[Enc(256)-Dec(256) gradient]{
  \includegraphics[width=45mm]{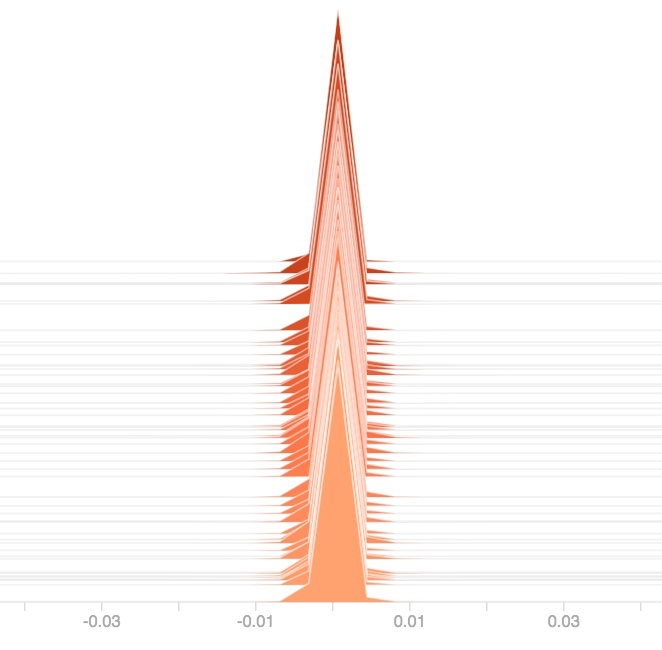}
}
\subfloat[Enc(1024)-Dec(256) gradient]{
  \includegraphics[width=45mm]{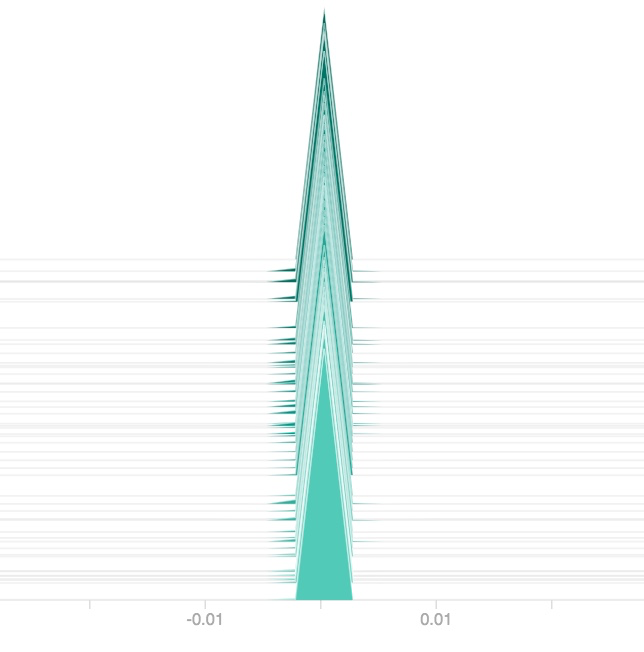}
}
\caption{Distribution of weight parameters and gradients of the sentence encoder.}
\label{fig:senenc_dist}
\end{figure*}

Figure~\ref{fig:docenc_dist} and \ref{fig:senenc_dist} present the distributions of model parameters and gradients of fully trainable models with different capacities. Note that models with different capacities have different scale of distribution. Models with smaller encoder hidden size tend to have larger scale of parameter and gradient values.

\end{document}